# An Overview of Indian Language Datasets Used for Text Summarization

Shagun Sinha[1] and Girish Nath Jha[2]

[1][2] Jawaharlal Nehru University, New Delhi, India
{[1]shagunsinha5, [2]girishjha}@gmail.com

**Abstract.** In this paper, we survey Text Summarization (TS) datasets in Indian Languages (ILs), which are also low-resource languages (LRLs). We seek to answer one primary question – is the pool of Indian Language Text Summarization (ILTS) dataset growing or is there a serious resource poverty? To answer the primary question, we pose two sub-questions that we seek about ILTS datasets - first, what characteristics – format and domain do ILTS datasets have? Second, how different are those characteristics of ILTS datasets from high-resource languages (HRLs) particularly English.

The survey of ILTS and English datasets reveals two similarities and one contrast. The two similarities are – first, the domain of dataset commonly is news (Hermann et al., 2015). The second similarity is the format of the dataset which is both extractive and abstractive. The contrast is in how the research in dataset development has progressed. ILs face a slow speed of development and public release of datasets as compared with English. We conclude that the relatively lower number of ILTS datasets is because of two reasons- first, absence of a dedicated forum for developing TS tools; and second, lack of shareable standard datasets in the public domain.

## 1 Introduction

In this paper, we present an overview of Text Summarization (TS) datasets available in Indian languages (ILs). TS is a sub-domain of Natural Language Processing (NLP) which is a field that lies at the intersection of computer science and linguistics. NLP develops computational "models and algorithms that enable a computer to interpret, process, and generate natural language" utterances (Lusetti et al., 2018, p. 1). Text Summarization (TS) is an NLP task in which computers are expected to summarize long texts. Thus, a TS tool produces summaries of input texts. The summary so produced can be one of the two types – first, extractive summary in which sentences from the source text are quoted verbatim; second, abstractive summary which is composed by rephrasing contents of the source text. Abstractive summaries are more coherent and human-like than extractive summaries due to which developing abstractive summarizers is tough (Kouris et al., 2021, p. 814).

Since the first work in TS by Luhn (1958), TS research has gradually gained attention. Rise of Machine Learning and Deep Learning methods has further enhanced the growth of TS research (Chen & Zhuge, 2018; Rush et al., 2015). Two key factors influence TS development– algorithm, and language for which TS is needed. The algorithm requires dataset of a given language for summarization training. TS datasets, like datasets for other NLP tasks, are easier to build for high-resource Languages (HRLs) than for low-resource languages (LRLs). Indian languages (ILs) are an example of LRLs because they lack sufficient digital content.

Most TS review articles review TS approaches and algorithms but do not specifically focus on the availability of datasets for ILs. (Hasan et al., 2021) observes that many languages lack publicly available datasets (p. 4700) . In this paper, we present a survey of TS datasets in English, an HRL, and ILs on the LRL front – Hindi, Malayalam, Sanskrit, Urdu, Kannada, Konkani, Punjabi, Bangla –with one primary objective of comparing their characteristics and pace of dataset growth. To assess the growth of TS efforts in ILs, it is important to review not just the algorithms and approaches to TS but also the availability of datasets that have been developed.

In other words, we present this survey with to compare ILTS datasets with the HRL English datasets. Through the comparison, we seek to answer two key questions – is the ILTS dataset pool growing or is it marred by resource poverty?

## 2    TS Datasets and Resources in English

Text Summarization has been the focus of many conferences which evolved out of the need to discuss better document- and information-processing technologies. Document, or text, summarization has been a regular part of these conferences organized by three US-based agencies NIST, DARPA and ARDA[1]. Some top conferences that evolved in the US in the early 2000s include[2] Document Understanding Conferences, Translingual Information Detection Extraction and Summarization (TIDES) conference, Text Retrieval Conference (TREC).

A few other groups that evolved as a language resource forum like the Linguistic Data Consortium (LDC) which started in 1992 at the University of Pennsylvania, Language Resource and Evaluation Conference (LREC) conference commenced to encourage research in language resources and development, Annual conference of the Association for Computational Linguistics (ACL), Conferences of the Association for Computing Machinery (ACM), Neural Image Processing (NIPS), and others. Some of the key summarization papers that emerged from NLP conferences include (Chen & Bansal, 2018), (Cabral et al., 2014), (Liu & Liu, 2009), (Sharma et al., 2019a), (Woodsend & Lapata, 2010). A detailed analysis of those conferences is out of the scope of this work.

---

[1] https://www-nlpir.nist.gov/projects/duc/intro.html
[2] https://www-nlpir.nist.gov/projects/duc/intro.html

### 2.1 Datasets

Some widely used training datasets in English include, XSum, Newsroom, Multi-News, CNN/DailyMail, MLSUM, and DUC datasets. Narayan et al. (2018) introduced XSum, a summarization dataset with 200K news articles in English with single line summary of each . Scialom et al. (2020) present a "multilingual extension" of the CNN/DailyMail dataset in French, German, Spanish, Russian, and Turkish, called MLSUM (p.8053).

Rush et al. (2015) developed a sentence summarization system using the news-headline pair of the Gigaword[3] dataset (p. 385) and evaluated it using the DUC-2003 and DUC-2004 datasets (p. 384).The DUC datasets have human generated summaries of news articles (p. 385) due to which they may be termed abstractive. Fabbri et al. (2019) present Multi-News which is a multi-document summarization dataset with summaries written by human editors (p. np). Newsroom[4] (Grusky et al., 2018) is a collection of article-summary pairs of 38 news publishers[5]. Similar is, CNN/DailyMail by Hermann et al. (2015). An exception to the news domain data includes the BIGPATENT dataset by (Sharma et al., 2019b) which is drawn from US Patent records(p.2204).

Datasets from the Data Understanding Conferences (DUC) have been regularly used in summarization evaluation tasks (D'Silva & Sharma, 2019) (p.1814). Similarly, as Jin et al. (2020) note the CNN DailyMail dataset is widely used in summarization evaluation (pp. 2000) although the dataset is not restricted to just evaluation and has been widely used in training and testing too like in Mishra and Gayen (2018), Kouris et al. (2021). A detailed account of key summarization datasets can be accessed in (Jin et al., 2020, p. 2000) and Dernoncourt et al. (2018).[6]

As Sharma et al. (2019b) argues, two factors may be observed in these datasets – first, news remains a widely used domain of TS dataset; second, these datasets are mostly extractive . We believe news articles remains a popular choice for developing TS dataset because news articles and their headlines provide a rough document-summary pattern making it easier to obtain a dataset rather than having to manually summarize articles.

## 3   TS Resources and Datasets in Indian Languages

NLP research in Indian Languages (ILs) is being undertaken by various institutions. While no standardized conferences exist for IL Text Summarization (TS), some conferences and individual researchers have made an attempt to build resources in both extractive and abstractive spheres for ILs. One characteristic of TS dataset is document-summary format of parallel data which is a suitable format for many supervised machine learning algorithms as opposed to monolingual corpus which are more suitable

---

[3] https://www.ldc.upenn.edu/ accessed January 20, 2022 at 21.12 hours

[4] Link : Papers with Code dataset page accessed January 25, 2022 at 21.09 hours

[5] https://paperswithcode.com/dataset/newsroom

[6] https://docs.google.com/spreadsheets/d/1b1-NpM1jDK7KVHd_CwrxhpNZ1zAE8m-7M0pZ0gfZTMQ/edit#gid=0  accessed January 10, 2022 at 23.21 hours

for unsupervised methods. We survey such parallel datasets in the 22 scheduled languages of the Indian constitution for the last 10 years (2012-2022). However, if a language does not have any parallel TS corpus, we cite the reported datasets in any given format. For every TS research work, we observe if the authors report the details of the dataset used, as well as if the dataset is currently accessible. We could also not find any TS research in 6 scheduled languages Bodo, Kashmiri, Manipuri, Maithili, Santhali, and Sindhi. Hence, those 6 languages have been excluded.

### 3.1    IL Datasets

**Assamese and Bengali.** Assamese TS by (Kalita et al., 2012) uses WordNet database for summarization (p.150) and thus, uses no text corpus. Talukder et al. (2019) present an abstractive summarization for Bengal using dataset from news corpus and social media data  (p. np) but provide no links to the dataset. Masum et al. (2019) develop a Bengali corpus for generating sentence similarity for abstractive summarization (p. np) but the corpus is not accessible. In Chowdhury et al. (2021) authors report using document-summary pairs from printed NCTB books (p. np) to develop an abstractive TS dataset but do not release the data for the research community.

**Hindi.** Two sources of Hindi TS are iNLTK and CDAC. iNLTK offers publicly available 'large and short corpora' for Hindi TS. CDAC hosts Saaranshak, an ontology-based multi-document summarizer that it developed[7]. However, Saaranshak is based on ontology and neither the tool nor the dataset is publicly available. Giri et al. (2016) observe that a TS for Hindi was proposed by Garain et al. (2006) and later adapted by CDAC (p. 54). But no details are present on the CDAC website to ascertain it is the same tool as authors argue.

**Malayalam.** Kabeer and Idicula (2014) develop Malayalam TS methods on news articles paired with human summaries (p. 146) and report their work as the first Malayalam TS (p. 150). Nambiar et al. (2021) report an abstractive TS on a BBC corpus that was translated in Malayalam (p. 351). Kishore et al. (2016) present a Malayalam TS method based on Paninian Grammar (p.197). None of these provides links to the datasets they used.

**Punjabi TS**. Jain et al. (2021) use Particle-Swarm Optimization for Punjabi Text Summarization using two corpora, a monolingual Punjabi corpus, and a Punjabi-Hindi parallel dataset from the Indian Language Corpora Initiative (p. 12). Similarly, Gupta and Kaur (2016) test their extractive methods on 150 random documents from two corpora – Punjabi text obtained by translating Hindi corpora by CILT, IIT Bombay and a Punjabi news corpus (p. 267). Other Punjabi TS works by Gupta and Lehal (2012) on news

---

dataset (p.200) and (Gupta & Lehal, 2013) provide no links to their datasets. Only the Punjabi monolingual corpora is available on TDIL[8].

**Urdu.** Humayoun et al. (2016) present an abstractive summarization corpus, Urdu Corpus (UC), built from 50 articles taken majorly from news articles and blogs (pp. 796-797). They provide datasets link. (Nawaz et al., 2020) present ETS models for Urdu for which they use the abstractive summary corpus provided by Humayoun et al. (2016) but they also develop an extractive version of the UC corpus, Urdu Corpus expert ground truth (p. 9) and Urdu Training Dataset obtained from some news sources (p. 10). The UC corpus is claimed to be open-access.

**Kannada and Konkani.** Kannada has witnessed many TS works. Shilpa and DR (2019) present a Kannada summarizer Abs-Sum-Kan which uses POS- and NER-techniques for summarizing documents (p. 1031). The authors do not mention any details about the dataset except a line about Kannada gazetteers compiled by Saha et al. (2008) (p. 1031). An earlier extractive or information retrieval (IR)-based approach to Kannada summarization by (Embar et al., 2013) does not provide any datasets. Guided summarization in Kannada (Kallimani et al., 2014) offers no datasets either. Konkani Literature-based TS dataset is offered by D'Silva and Sharma (2019) and it has 71 stories (p. 1814). While the dataset is limited in number, the domain of folk tales is unique.

**Sanskrit, Dogri, and Oriya.** The sole experiment in Sanskrit TS so far is Barve et al. (2016) which is an extractive TS on Sanskrit Wikipedia articles. Authors Barve et al. (2016) do not provide any dataset links but Sanskrit Wikipedia articles have been presented in (Arora, 2020). Dogri TS work by (Gandotra & Arora, 2021) is not accessible and the paper metadata does not provide any details about the data. In (Pattnaik & Nayak, 2020) authors report a manually built dataset of 200 cricket news articles with human summaries (p. 826) and in (Pattnaik & Nayak, 2021), authors report a news articles-summary texts used as input to their proposed system (p. 397), but the authors do not release those datasets.

**Gujarati, Marathi, and Nepali.** Sarwadnya and Sonawane (2018) use Marathi news articles and Marathi portion of EMILLE[9] dataset (p.np). Ranchhodbhai (2016) also evaluates his extractive summarization approach on Gujarati EMILLE corpus (p.117). (Khanal et al., 2021) collect Nepali news articles to implement extractive methods (p. 989) but the dataset is not released.

**Telugu and Tamil.** Rani et al. (2021) use a Telugu corpus provided openly accessible on Kaggle[10] (p.358). (Naidu et al., 2018) uses e-newspaper data of Telugu (p.562) but

---

do not release it. (Mohan Bharath et al., 2022) use manually created dataset as per their abstract but the paper is not accessible and hence, no further details are available. (Priyadharshan & Sumathipala, 2018) use Tamil sports news as the corpus for TS evaluation. Their paper is not openly accessible to gather further details.

**A good development in ILTS** is translated and publicly accessible TS dataset. A multilingual corpus from PIB and Mann ki Baat data has been presented in Siripragada et al. (2020). offers TS dataset derived from BBC in 44 languages including a few major ILs like Bengali. XL-SUM is claimed to be the first public dataset in abstractive summarization for many languages (p. 4700). However, Arora (2020) had released Hindi summarization corpus prior to their work.

## 4    Observations

We observe two similarities and one contrast between English and ILTS datasets. First, as noted earlier, news articles are the common source of TS datasets in English. A similar pattern has been observed in ILTS, as seen above. ILTS datasets derive from news articles with a few exceptions like Konkani folktales dataset (D'Silva & Sharma, 2019). Second, the reported datasets have been used in extractive as well as abstractive TS development.

We observe a stark contrast in the dataset development efforts though. TS datasets remain largely difficult to be found. Datasets like the CNN/Dailymail or the DUC datasets in English, on the other hand, are not only open access, they are also widely adapted. There are two possible reasons for such limited reach of ILTS datasets - first, ILTS has not been the focus of any Indic NLP groups in India. We noted a few forums for English TS like the DUC, LDC, TREC which have had dedicated tracks for summarization for years. However, ILTS does not have any dedicated platforms. While TDIL provides access to monolingual and parallel corpora which are usually used for machine translation, a dedicated platform for TS is still missing; second, lack of open access datasets. As can be seen in Table1 and in the previous section, many works report datasets but never release it for the research community. Hence, datasets are used and quoted but not made open access for the research community thereby impeding research progress in TS. To the best of our knowledge, these reasons have not been observed as the cause of data paucity in ILTS so far.

Two exceptions to limited dataset availability include - first, the set of online platforms like HuggingFace[11], iNLTK[12], Tensorflow[13], nlpprogress[14], and paperswithcode[15] which provide open access datasets across languages including ILs. A detailed analysis of those platforms is out of the scope of this work but those platforms include

---

many languages including some ILs. A dedicated TS platform for ILs is still missing; second, in some cases of extractive summarization, researchers use monolingual corpus which may be open access but such corpora are general-purpose which have not been developed for TS specifically. Thus, TS-focused dataset, especially in the form of text-summary pairs, are difficult to obtain.

## 5 Conclusion

Language datasets are important for training NLP algorithms including TS algorithms. ILs are LRLs due to which dataset development for TS requires special attention. In this paper, we surveyed TS datasets as reported by various researchers and developers to find out two similarities and one contrast of ILTS with datasets in English, which is an HRL. Similarities are in the domain and approach of summarization. Like English datasets, many ILTS datasets derive from news sources and are used in both extractive and abstractive summarization tasks.

The contrast, however, is in the accessibility of datasets. Unlike English datasets, most IL datasets are limited in availability and are not open access. More TS-focused efforts and research forums could make more datasets available.

**Table 1. Text Summarization Datasets in Indian Languages**

| Language | Dataset used | Source | TS Approach used | Open Access? Y/N |
|---|---|---|---|---|
| Malayalam | 1. Malayalam news articles | (Kabeer & Idicula, 2014) | Extractive & Abstractive | No |
| | 2.Translated BBC corpus | (Nambiar et al., 2021) | Abstractive | No |
| Kannada | Gazetteer data by (Saha et al., 2008) | Shilpa and DR (2019) | Abstractive | No |
| Marathi | EMILLE dataset by Lancester Univ. | (Sarwadnya & Sonawane, 2018) | Extractive | NA |
| Assamese | WordNet database | (Kalita et al., 2012) | Extractive | NA |
| Konkani | Folktales Dataset | (D'Silva & Sharma, 2019) | Abstractive | No |
| Telugu | News corpus from Kaggle | (Rani et al.) | Extractive | **NA** |
| | Telugu dataset | (Mohan Bharath et al., 2022) | Abstractive | **NA** |
| Tamil | e-News corpus | (Priyadharshan & Sumathipala, 2018) | Extractive | e-Newspaper could be accessible |
| Bengali | NCTB book dataset | (Chowdhury et al., 2021) | Abstractive | No |
| Urdu | 1. Urdu Summary Corpus (UC) | (Humayoun et al., 2016) | Abstractive | Claimed to be open access |
| | 2.UCE and UTD | (Nawaz et al., 2020) | Extractive | NA |
| Sanskrit | Wikipedia articles | (Barve et al., 2016) | Extractive | Available in (Arora, 2020) |
| Gujarati | EMILLE corpus for evaluation | (Ranchhodbhai, 2016) | Extractive | NA |
| Nepali | Nepali News corpus | (Khanal et al., 2021) | Extractive | No |
| Oriya | News corpus with human summaries | (Patnaik & Nayak, 2020), (2021) | NA | No |
| Hindi | Hindi news articles | (Arora, 2020) | Abstractive | Yes |
| Punjabi | ILCI monolingual corpus from TDIL | (Jain et al., 2021) | Extractive | Monolingual corpus available on request |

Dogri          NA          (Gandotra & Arora, 2021)          NA                    NA

## References

Arora, G. (2020). iNLTK: Natural Language Toolkit for Indic Languages. Proceedings of Second Workshop for NLP Open Source Software (NLP-OSS),

Barve, S., Desai, S., & Sardinha, R. (2016). Query-based extractive text summarization for Sanskrit. Proceedings of the 4th International Conference on Frontiers in Intelligent Computing: Theory and Applications (FICTA) 2015,

Cabral, L. d. S., Lins, R. D., Mello, R. F., Freitas, F., Ávila, B., Simske, S., & Riss, M. (2014). *A platform for language independent summarization* Proceedings of the 2014 ACM symposium on Document engineering, Fort Collins, Colorado, USA. https://doi.org/10.1145/2644866.2644890

Chen, J., & Zhuge, H. (2018). Abstractive text-image summarization using multi-modal attentional hierarchical rnn. Proceedings of the 2018 Conference on Empirical Methods in Natural Language Processing,

Chen, Y.-C., & Bansal, M. (2018). Fast abstractive summarization with reinforce-selected sentence rewriting. *arXiv preprint arXiv:1805.11080*.

Chowdhury, R. R., Nayeem, M. T., Mim, T. T., Chowdhury, M., Rahman, S., & Jannat, T. (2021). Unsupervised Abstractive Summarization of Bengali Text Documents. *arXiv preprint arXiv:2102.04490*.

D'Silva, J., & Sharma, U. (2019). Development of a Konkani Language Dataset for Automatic Text Summarization and its Challenges. *International Journal of Engineering Research and Technology. International Research Publication House. ISSN*, 0974-3154.

Dernoncourt, F., Ghassemi, M., & Chang, W. (2018). A repository of corpora for summarization. Proceedings of the Eleventh International Conference on Language Resources and Evaluation (LREC 2018),

Embar, V. R., Deshpande, S. R., Vaishnavi, A., Jain, V., & Kallimani, J. S. (2013). sArAmsha-A Kannada abstractive

summarizer. 2013 International Conference on Advances in Computing, Communications and Informatics (ICACCI),

Fabbri, A. R., Li, I., She, T., Li, S., & Radev, D. R. (2019). Multi-news: A large-scale multi-document summarization dataset and abstractive hierarchical model. *arXiv preprint arXiv:1906.01749*.

Gandotra, S., & Arora, B. (2021). Feature Selection and Extraction for Dogri Text Summarization. In *Rising Threats in Expert Applications and Solutions* (pp. 549-556). Springer.

Garain, U., Datta, A. K., Bhattacharya, U., & Parui, S. K. (2006). Summarization of jbig2 compressed indian language textual images. 18th International Conference on Pattern Recognition (ICPR'06),

Giri, V. V., Math, M., & Kulkarni, U. (2016). A survey of automatic text summarization system for different regional language in India. *Bonfring International Journal of Software Engineering and Soft Computing*, *6*(Special Issue Special Issue on Advances in Computer Science and Engineering and Workshop on Big Data Analytics Editors: Dr. SB Kulkarni, Dr. UP Kulkarni, Dr. SM Joshi and JV Vadavi), 52-57.

Grusky, M., Naaman, M., & Artzi, Y. (2018). Newsroom: A Dataset of 1.3 Million Summaries with Diverse Extractive Strategies. Proceedings of the 2018 Conference of the North American Chapter of the Association for Computational Linguistics: Human Language Technologies, Volume 1 (Long Papers),

Gupta, V., & Kaur, N. (2016). A novel hybrid text summarization system for Punjabi text. *Cognitive Computation*, *8*(2), 261-277.

Gupta, V., & Lehal, G. S. (2012). Complete pre processing phase of Punjabi text extractive summarization system. Proceedings of COLING 2012: Demonstration Papers,

Gupta, V., & Lehal, G. S. (2013). Automatic text summarization system for Punjabi language. *journal of emerging technologies in web intelligence*, *5*(3), 257-271.

Hasan, T., Bhattacharjee, A., Islam, M. S., Mubasshir, K., Li, Y.-F., Kang, Y.-B., Rahman, M. S., & Shahriyar, R. (2021). XL-Sum: Large-Scale Multilingual Abstractive Summarization for 44 Languages. Findings of the Association for Computational Linguistics: ACL-IJCNLP 2021,

Hermann, K. M., Kocisky, T., Grefenstette, E., Espeholt, L., Kay, W., Suleyman, M., & Blunsom, P. (2015). Teaching machines to read and comprehend. *Advances in neural information processing systems*, *28*, 1693-1701.

Humayoun, M., Nawab, R. M. A., Uzair, M., Aslam, S., & Farzand, O. (2016). Urdu summary corpus. Proceedings of the Tenth International Conference on Language Resources and Evaluation (LREC'16),

Jain, A., Yadav, D., & Arora, A. (2021). Particle Swarm Optimization for Punjabi Text Summarization. *International Journal of Operations Research and Information Systems (IJORIS)*, *12*(3), 1-17.

Jin, H., Cao, Y., Wang, T., Xing, X., & Wan, X. (2020). Recent advances of neural text generation: Core tasks, datasets, models and challenges. *Science China Technological Sciences*, 1-21.

Kabeer, R., & Idicula, S. M. (2014). Text summarization for Malayalam documents—An experience. 2014 International Conference on Data Science & Engineering (ICDSE),

Kalita, C., Saharia, N., & Sharma, U. (2012). An extractive approach of text summarization of assamese using wordnet. Global WordNet Conference (GWC-12),

Kallimani, J. S., Srinivasa, K., & Reddy, B. E. (2014). A Comprehensive Analysis of Guided Abstractive Text Summarization. *International Journal of Computer Science Issues (IJCSI)*, *11*(6), 115.

Khanal, R. S., Adhikari, S., & Thapa, S. (2021). Extractive Method for Nepali Text Summarization Using Text Ranking and LSTM. 10th IOE Graduate Conference,

Kishore, K., Gopal, G. N., & Neethu, P. (2016). Document Summarization in Malayalam with sentence framing. 2016 International Conference on Information Science (ICIS),

Kouris, P., Alexandridis, G., & Stafylopatis, A. (2021). Abstractive Text Summarization: Enhancing Sequence-to-Sequence Models Using Word Sense Disambiguation and Semantic Content Generalization. *Computational Linguistics*, *47*(4), 813-859. https://doi.org/10.1162/coli_a_00417

Liu, F., & Liu, Y. (2009). From extractive to abstractive meeting summaries: Can it be done by sentence compression?

Proceedings of the ACL-IJCNLP 2009 Conference Short Papers,

Luhn, H. P. (1958). The automatic creation of literature abstracts. *IBM Journal of research and development*, *2*(2), 159-165.

Lusetti, M., Ruzsics, T., Göhring, A., Samardžić, T., & Stark, E. (2018). Encoder-decoder methods for text normalization.

Masum, A. K. M., Abujar, S., Tusher, R. T. H., Faisal, F., & Hossain, S. A. (2019). Sentence similarity measurement for Bengali abstractive text summarization. 2019 10th International Conference on Computing, Communication and Networking Technologies (ICCCNT),

Mishra, R., & Gayen, T. (2018). Automatic Lossless-Summarization of News Articles with Abstract Meaning Representation. *Procedia Computer Science*, *135*, 178-185.

Mohan Bharath, B., Aravindh Gowtham, B., & Akhil, M. (2022). Neural Abstractive Text Summarizer for Telugu Language. In *Soft Computing and Signal Processing* (pp. 61-70). Springer.

Naidu, R., Bharti, S. K., Babu, K. S., & Mohapatra, R. K. (2018). Text summarization with automatic keyword extraction in telugu e-newspapers. In *Smart computing and informatics* (pp. 555-564). Springer.

Nambiar, S. K., Peter, S. D., & Idicula, S. M. (2021). Abstractive Summarization of Malayalam Document using Sequence to Sequence Model. 2021 7th International Conference on Advanced Computing and Communication Systems (ICACCS),

Narayan, S., Cohen, S. B., & Lapata, M. (2018). Don't give me the details, just the summary! topic-aware convolutional neural networks for extreme summarization. *arXiv preprint arXiv:1808.08745*.

Nawaz, A., Bakhtyar, M., Baber, J., Ullah, I., Noor, W., & Basit, A. (2020). Extractive Text Summarization Models for Urdu Language. *Information Processing & Management*, *57*(6), 102383. https://doi.org/https://doi.org/10.1016/j.ipm.2020.102383

Pattnaik, S., & Nayak, A. K. (2020). A simple and efficient text summarization model for odia text documents. *Indian journal of computer science and engineering*, *11*(6), 825-834. https://doi.org/ 10.21817/indjcse/2020/v11i6/201106132

Pattnaik, S., & Nayak, A. K. (2021). Automatic Text Summarization for Odia Language: A Novel Approach. In *Intelligent and Cloud Computing* (pp. 395-403). Springer.

Priyadharshan, T., & Sumathipala, S. (2018). Text summarization for Tamil online sports news using NLP. 2018 3rd international conference on information technology research (ICITR),

Ranchhodbhai, S. J. (2016). *Designing and Development of Stemmer and String Similarity Measure for Gujarati Language and their Application in Text Summarization System*

Rani, B. K., Rao, M. V., Srinivas, K., & Madhukar, G. (2021). Telugu Text Summarization using LSTM Deep Learning. *Pensee Journal Vol*, *51*(1), 355-363.

Rush, A. M., Chopra, S., & Weston, J. (2015). A Neural Attention Model for Abstractive Sentence Summarization. Proceedings of the 2015 Conference on Empirical Methods in Natural Language Processing,

Saha, S. K., Sarkar, S., & Mitra, P. (2008). Gazetteer preparation for named entity recognition in indian languages. Proceedings of the 6th Workshop on Asian Language Resources,

Sarwadnya, V. V., & Sonawane, S. S. (2018). Marathi extractive text summarizer using graph based model. 2018 Fourth International Conference on Computing Communication Control and Automation (ICCUBEA),

Scialom, T., Dray, P.-A., Lamprier, S., Piwowarski, B., & Staiano, J. (2020). MLSUM: The Multilingual Summarization Corpus. Proceedings of the 2020 Conference on Empirical Methods in Natural Language Processing (EMNLP),

Sharma, E., Li, C., & Wang, L. (2019a). Bigpatent: A large-scale dataset for abstractive and coherent summarization. *arXiv preprint arXiv:1906.03741*.

Sharma, E., Li, C., & Wang, L. (2019b). BIGPATENT: A Large-Scale Dataset for Abstractive and Coherent Summarization. Proceedings of the 57th Annual Meeting of the Association for Computational Linguistics,

Shilpa, G., & DR, S. K. (2019). Abs-Sum-Kan: An Abstractive Text Summarization Technique for an Indian Regional Language by Induction of Tagging Rules. *International Journal of Recent*

*Technology and Engineering*, *8*(2S3), 1028-1036. https://doi.org/10.35940/ijrte.B1193.0782S319

Siripragada, S., Philip, J., Namboodiri, V. P., & Jawahar, C. (2020). A Multilingual Parallel Corpora Collection Effort for Indian Languages. Proceedings of the 12th Language Resources and Evaluation Conference,

Talukder, M. A. I., Abujar, S., Masum, A. K. M., Faisal, F., & Hossain, S. A. (2019). Bengali abstractive text summarization using sequence to sequence RNNs. 2019 10th International Conference on Computing, Communication and Networking Technologies (ICCCNT),

Woodsend, K., & Lapata, M. (2010). Automatic generation of story highlights. Proceedings of the 48th Annual Meeting of the Association for Computational Linguistics,